\documentclass[conference]{IEEEtran}
\IEEEoverridecommandlockouts
\usepackage{cite}
\usepackage{amsmath,amssymb,amsfonts}
\usepackage{algorithmic}
\usepackage{graphicx}
\usepackage{textcomp}
\usepackage{xcolor}
\usepackage{url}
\def\BibTeX{{\rm B\kern-.05em{\sc i\kern-.025em b}\kern-.08em
    T\kern-.1667em\lower.7ex\hbox{E}\kern-.125emX}}
\begin{document}

\title{Self-supervised Activity Representation Learning with Incremental Data: An Empirical Study
%\thanks{* equal contribution}
}

\author{\IEEEauthorblockN{Jason Liu\textsuperscript{\textsection},  
Shohreh Deldari\textsuperscript{\textsection}, 
Hao Xue \textsuperscript{\textsection},
Van Nguyen\textsuperscript{\textsection\textsection},
Flora D. Salim\textsuperscript{\textsection}}
\IEEEauthorblockA{
\textsuperscript{\textsection} School of computer science and engineering,
University of New South Wales (UNSW), Australia\\
\textsuperscript{\textsection\textsection} Defence Science and Technology Group, Australia\\
Email: jason.liu3@student.unsw.edu.au,  \{s.deldari, hao.xue1, flora.salim\}@unsw.edu.au} and van.nguyen5@defence.gov.au
}

\maketitle

\begin{abstract}
In the context of mobile sensing environments, various sensors on mobile devices continually generate a vast amount of data. Analyzing this ever-increasing data presents several challenges, including limited access to annotated data and a constantly changing environment. Recent advancements in self-supervised learning have been utilized as a pre-training step to enhance the performance of conventional supervised models to address the absence of labelled datasets. This research examines the impact of using a self-supervised representation learning model for time series classification tasks in which data is incrementally available.
We proposed and evaluated a workflow in which a model learns to extract informative features using a corpus of unlabeled time series data and then conducts classification on labelled data using features extracted by the model. We analyzed the effect of varying the size, distribution, and source of the unlabeled data on the final classification performance across four public datasets, including various types of sensors in diverse applications.
\end{abstract}

\begin{IEEEkeywords}
self-supervised representation learning, incremental learning,  mobile sensing, activity recognition
\end{IEEEkeywords}

%\begingroup\renewcommand\thefootnote{*}
%\footnotetext{Equal contribution.}

\section{Introduction}

In real-world situations, obtaining a significant amount of labelled data takes considerable time and effort due to various issues such as privacy concerns, lack of domain knowledge, or time and budget constraints. Consequently, the conventional supervised learning-based AI methods are not suitable. Recently, \textit{self-supervised learning} (SSL) techniques have been developed that can generalize to various tasks, data domains, and input structures \cite{ts2vec,tnc,franceschi}. Self-supervised representation learning (SSRL) as an SSL technique is often done using unlabeled data by constructing pair of samples and corresponding pseudo-labels based on the original data points and training the model upon that to learn the structure of the data. Existing research has focused on using as few as possible of the labelled data to achieve performance comparable with that of using all labelled data ~\cite{tstcc,gorade}. 

Although self-supervised learning for sensors has been recently investigated for various applications \cite{Deldari_2022,deldari_2021,haresamudram2020masked,saeed2021sense,xue2021exploring,ts2vec,tstcc}, this has yet to be done for situations where all data is not available at once. Hence, dynamic changes in the incoming data (e.g., encountering new tasks and classes, change of distributions, etc.) could cause the model to lose its optimality. Incremental learning refers to the ability of a machine learning model to learn and adapt to new information over time. This %is a difficult problem to solve because it 
requires the model to retain past knowledge while also being able to incorporate new information and make predictions about previously unseen data. 
We hypothesized that a self-supervised representation learning model could benefit from the extensive amount of available unlabeled data to improve its feature extraction capability and classification performance, similar to how supervised classifiers can use incremental data to improve performance on both existing and new tasks~\cite{gupta,oilfts}.

The main focus of this investigation is to determine if the pre-training phase with a self-supervised representation learning model improves with the availability of more training data. We consider several scenarios in order to answer this question. For example, we want to know if increasing the amount of data used in the pre-training phase leads to better classification accuracy downstream and whether there is a difference in accuracy when pre-training on the same dataset as the one being classified versus a dataset with similar measurements but collected in a different context.

Continual learning (CL) is about making use of data that becomes available over time, using constant computational and memory resources to develop more complex knowledge about the subject area incrementally. The main challenge in any CL framework is addressing the issue of catastrophic forgetting, which is the tendency of neural network models to forget previously learned information upon being fitted with new data. A fair number of studies on representation learning with continual learning explored various data types such as image, speech, and text. However, to the best of our knowledge, in terms of time series classification, there has yet to be any investigation on using incremental data to improve the self-supervised representation learning models used in the pre-training phase. This work explores the impact of SSL with incremental data on time series data (wearable sensors specifically).
The main contributions of this work are:

\begin{figure*}
    \centering
    \includegraphics[width=0.9\linewidth]{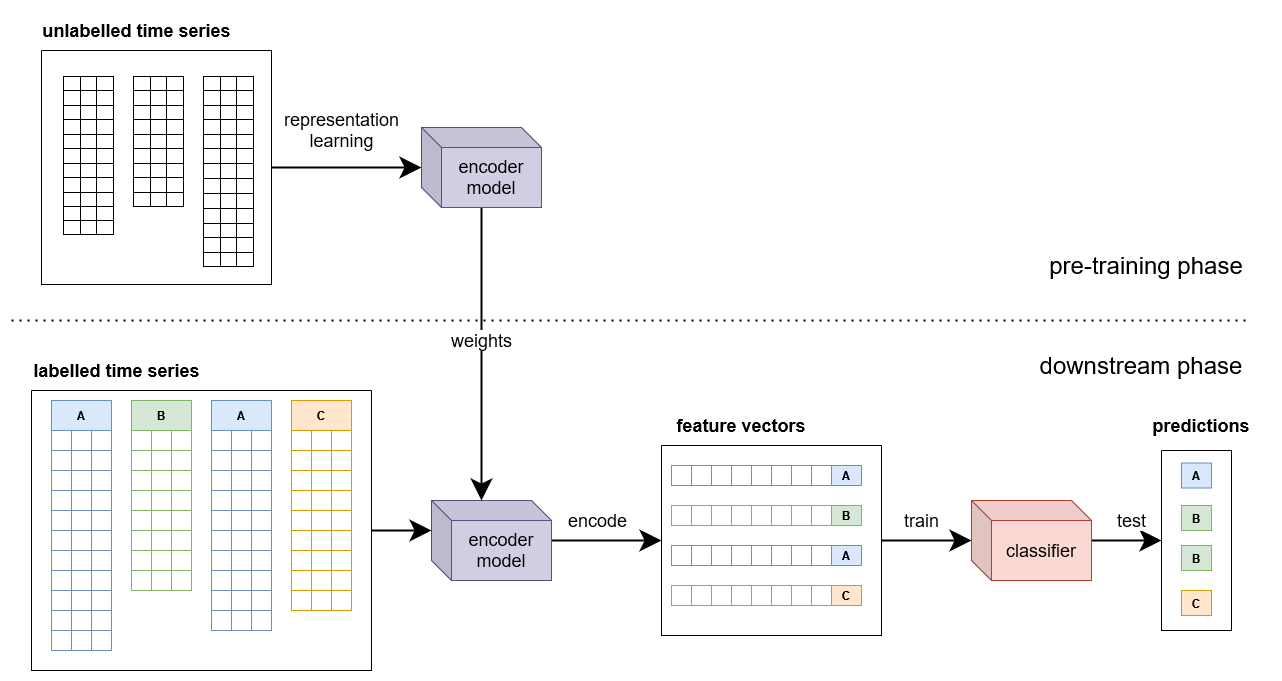}
    \caption{Overview of the workflow. Unlabelled time series data were used to train an encoder model in a pre-training phase, which is then used in a downstream phase to extract features from labelled data and served as input to the downstream classifier.}
    \label{fig:framework}
\end{figure*}

\begin{itemize}
    \item We investigated the contribution of self-supervised representation learning in incremental training setup across three well-known public human activity recognition datasets. We provided a training and evaluation framework of the improvements gained by the self-supervised representation learning model in both instance and class incremental continual learning context.
    \item We examined the effect of training the encoder and classifier using data from various domains and distributions.
    \item We evaluated the effectiveness of self-supervised representation learning in low-label data regimes.
    
\end{itemize}

\section{Research Gaps}
In the last few years, extensive work has been done on self-supervised representation learning with few-labelled data to retain comparable performance with their supervised counterpart in different domains such as computer vision \cite{chen2020simple,fini2022self,gomez2022continually}, natural language processing \cite{devlin2018bert}, mobile and ubiquitous sensor data \cite{Deldari_2022,deldari_2021, ts2vec, tnc}. \cite{deldari2022beyond} provides a comprehensive review of SSL methods across different modalities.%There has been a growing interest in using self-supervised learning within the context of continual learning for data types beyond time series, particularly in the field of computer vision \cite{}. 
However, the ability of self-supervised learning models to learn unseen data and improve the performance of existing continual learning frameworks to adapt and learn continuously from new and changing data is yet to be explored \cite{fini2022self}. 

Li et al. elucidate the benefit of continual representation learning both theoretically and empirically and sheds light on the role of the task order, diversity, and sample size. They also proposed a new CL algorithm ESPN~\cite{espn}, to learn quality representations on images. Extensive experimental evaluations demonstrate its ability to achieve good accuracy as well as fast inference.
Hsu et al. presented the first controlled study to better understand the domain shift in self-supervised learning for automatic speech recognition~\cite{hsu}. Results show that adding unlabelled in-domain data improves performance, even when the fine-tuning data does not match the test domain. Self-supervised representations trained on various domains are robust and lead to better generalisation performance on domains completely unseen during pre-training and fine-tuning. We will explore whether the findings of other data types also apply in the case of time series data.

\section{Method}
Our research aims to explore the potential of using large amounts of unlabeled time series data in a self-supervised manner to improve the classification accuracy of supervised models in a real-world situation where the input data and classes are provided incrementally. Specifically, we aim to determine whether a representation learning model can benefit from incrementally arriving unlabeled data and whether it is worthwhile to utilize continual learning (CL) strategies for this purpose. Our focus is on time series classification, and we are interested in understanding the impact that self-supervised representation learning has on this task. 

\subsection{Investigated Framework}
The proposed framework, as depicted in Figure \ref{fig:framework}, lies in the intersection of several research areas: we investigate the potential benefits of leveraging incremental data to a self-supervised representation learning model in the context of time series classification. In our study, we first pre-train the representation learning model (encoder) with unlabelled data of varying sizes, distributions, and domains. In this step, we employ \textit{TS2Vec} \cite{ts2vec}, one of the SOTA SSL models for time series data, as the encoder backbone. \textit{Pre-training} is the process of fitting the encoder model on unlabelled (or labels removed) data to learn its features. In the next step, the encoder is used as is or fine-tuned according to the final task, along with the downstream classifier. The main purpose of the encoder is to map the high-dimensional input data to a lower-dimension feature vector that can represent the class of input. To assess the impact of incremental pre-training, we maintain separate training sets for each phase.
We investigate the effectiveness of self-supervised pretraining in two different setups, including:
\begin{itemize}
    \item \textit{Instance incremental}: In this setting, the dataset is split into non-overlapping batches or subsets based on the subjects in each dataset, through random division.
    \item \textit{Class incremental}: in this setting, the dataset is partitioned by the class label of the instances into non-overlapping sets.
    %\item \textit{Low-labeled}:
\end{itemize}

\section{Empirical Studies}

\subsection{Datasets}
Our study utilized four widely used datasets for human activity recognition that have been heavily relied upon in previous works across time-series and wearable sensor research, as well as in the self-supervised representation learning domain's state-of-the-art models \cite{ts2vec,deldari_2021,Deldari_2022}.
\subsubsection{UEA \cite{uea}} UEA benchmark datasets \footnote{http://www.timeseriesclassification.com} consists of 30 datasets with a wide range of cases, dimensions, and series lengths. All are formatted to be of equal length, with no missing data and provided train/test splits. The datasets are grouped into the following categories: Human Activity Recognition (HAR), Motion classification, ECG classification, EEG/MEG classification, Audio Spectra Classification, and others.
%We used the cloned TS2Vec repository and followed the instructions for downloading and pre-processing the 30 UEA datasets, these were saved in their original formats~\cite{ts2vec}. These datasets were class-balanced, have uniform series length, no missing values, and were provided with a training and testing split.

\subsubsection{UCIHAR \cite{ucihar}} UCIHAR dataset \footnote{\url{https://archive.ics.uci.edu/ml/datasets/human+activity+recognition+using+smartphones}} is a human activity recognition dataset that contains recordings from 30 volunteers who carried out six activities including walking, walking upstairs, walking downstairs, sitting, standing, and lying. Activities are recorded by a smartphone device mounted on the volunteer’s waist. The data was in the format of the sliding window with 50\% overlap and already split into a train set of 21 subjects and a test set of 9 subjects. %The X, Y, and Z dimension of the body accelerometer and gyroscope features were taken from the raw inertial signals. The sliding windows were joined, with overlap removed. Both the train and test sets were combined and grouped by subject. For each subject, entries with missing values were dropped, then normalised. Subject data were further grouped by consecutive entries with the same activity, this defined a single activity instance from that subject. All instances from the same subject were saved together, with the corresponding activity labels saved separately.

\begin{figure}
    \centering
    \includegraphics[width=\linewidth]{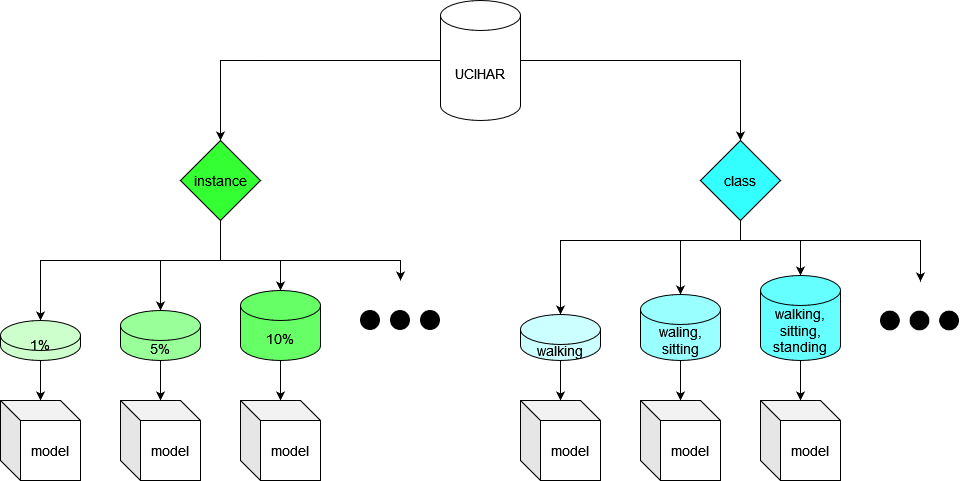}
    \caption{Incremental pre-train data settings for UCIHAR. In instance-incremental, models were trained for various proportions of instances. In class incremental, models were trained for various number of classes.}
    \label{fig:har_incr}
\end{figure}

\subsubsection{PAMAP2 ~\cite{pamap2}} Physical Activity Monitoring dataset \footnote{https://archive.ics.uci.edu/ml/datasets/pamap2+physical+activity+monitoring}, contains data of 18 different physical activities performed by nine subjects wearing three inertial measurement units and a heart rate monitor. In this set of experiments, we only used three accelerometers and gyroscope sensor data and 12 activities, including lie, sit, stand, iron, vacuum, ascend stairs, descend stairs, walk, Nordic walk, cycle, run, and rope jump. 

\subsubsection{Opportunity ~\cite{opportunity}} Opportunity dataset \footnote{http://opportunity-project.eu} consists of data collected from IMU sensor from 4 participants performing activities of daily living with 17 on-body sensor devices. We used the accelerometer and gyroscope from the back inertial measurement unit for our evaluation.

\subsection{Evaluation Setup}
The representation learning model is used to encode the raw time series instance into a representation. We pre-train the TS2Vec encoder with unlabelled time series instances, producing an encoder model that can be used to extract features from other instances. 
We used the HAR datasets (UCIHAR, PAMAP2, Opportunity) to see how an encoder model pre-trained on one dataset performs when tested on another dataset. We also measured the performance across various proportions and numbers of classes used during pre-training. 
For each HAR dataset, $1/4$ of the subjects were randomly chosen as the pre-training dataset (UCIHAR: 8, PAMAP2: 2, Opportunity: 1). An encoder model (TS2Vec) was pre-trained on the selected subjects for each of the datasets. The UCIHAR dataset was chosen in particular to investigate instance and class incremental continual learning settings, as illustrated in Figure~\ref{fig:har_incr}. Therefore, as depicted in this figure, we incrementally add new classes or new subjects to the training set and train the model based on that. To allow for more variation in the data, the original training split (21 subjects) was used for the pre-training phase, and each class of activity contains approximately an equal proportion of instances.

\begin{figure}
    \centering
    \includegraphics[width=\linewidth]{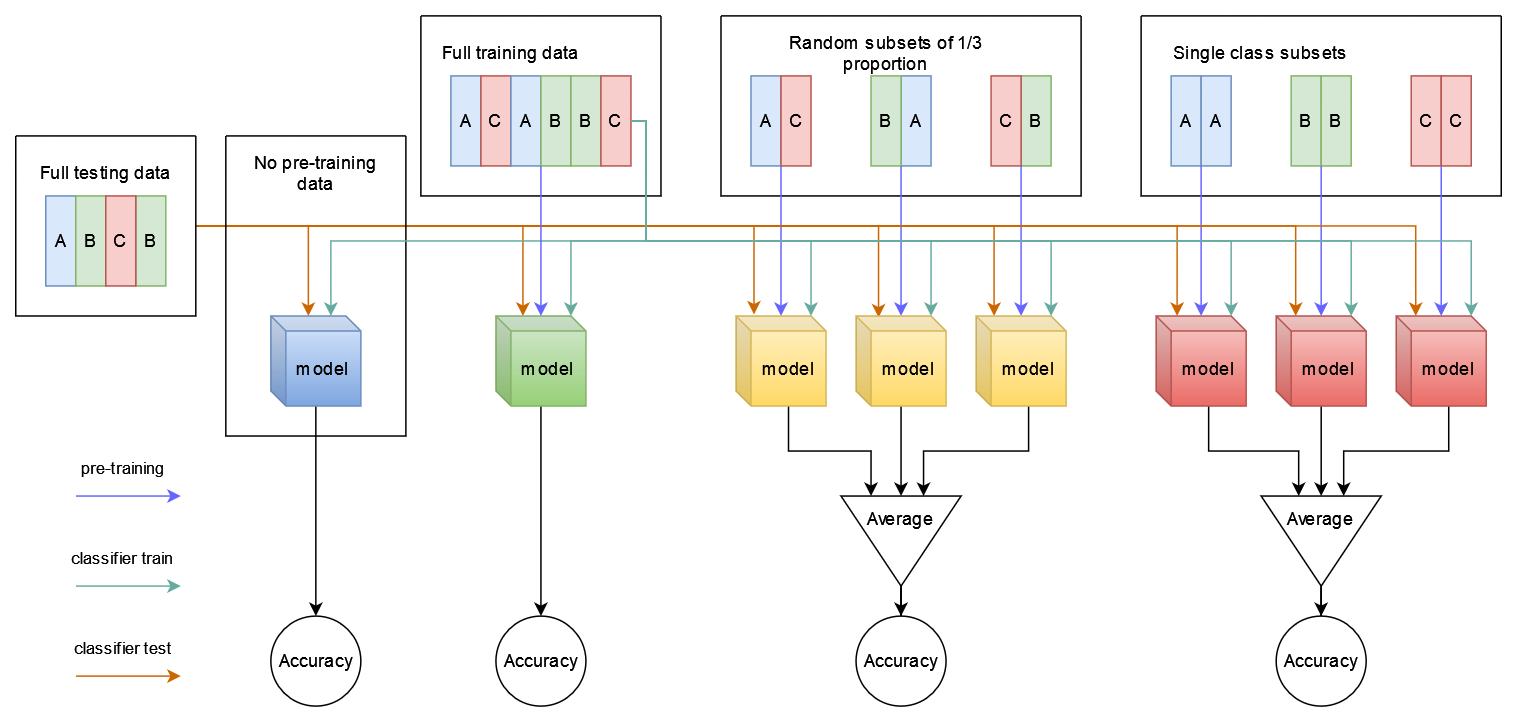}
    \caption{Workflow of the UEA dataset investigation. Full training data were split into random subsets with size according to the number of classes in a dataset, and also single class subsets. These were used to pre-train encoder models, and compared with model trained based on no and full pre-training data.}
    \label{fig:uea_workflow}
\end{figure}

We utilized Avalanche \cite{lomonaco2021avalanche}, a python-based library,  to attempt a naive instance and class incremental continual learning strategy on each datasets. Each dataset was converted to a stream of experiences, where each experience contains a subset of training data according to the incremental setting. For instance-incremental, there were a $1/m$ proportion of data in each experience where $m$ is the number of classes in that dataset; and for class incremental, a single class in each experience. Within each experience, a new encoder model was pre-trained on the data for that experience; the models from all experiences were aggregated to evaluate the performance in that continual learning setting. For each dataset, the train and test sets were normalised based on the train set. The entire train set was used to fit the downstream SVM classifier, and evaluations are based on the test set. %In settings where multiple pre-train models were used for different subsets of the data, the downstream classifier was trained on the whole train set and evaluated on the whole test set.Finally, the average accuracy obtained from each model is reported.
Figure~\ref{fig:uea_workflow} shows an overview of this workflow for the UEA dataset.
In all experiments, we compared the results against a randomly-initialized model with no pre-training phase and a fully supervised classifier as lower-bound and upper-bound baselines, respectively.

\subsection{Experiments}

\begin{table}
\caption{Accuracy achieved by different baselines w/o pretraining phase across three HAR datasets. Our proposed SSL-based framework consistently reaches high accuracy.}
\centering
\begin{tabular}{|l|c|c|c|c|}
\hline
Model              & Pretraining & UCIHAR & PAMAP2 & Opportunity \\ \hline
\hline
TS2Vec & None        & 0.92   & 0.74   & 0.87        \\ \hline
TS2Vec &UCIHAR      & \textbf{0.97}   & 0.81   & 0.86        \\ \hline
TS2Vec &PAMAP2      & 0.95   & 0.84   & 0.85        \\ \hline
TS2Vec &Opportunity & 0.93   & \textbf{0.89}   & 0.86        \\ \hline
InceptionTime &  -   & \textbf{0.97}   & 0.87   & \textbf{0.90}        \\ \hline
LSTM    &    -       & 0.20   & 0.16   & 0.46        \\ \hline
GRU   &    -         & 0.20   & 0.15   & 0.46        \\ \hline
TCN    &  -          & 0.60   & 0.16   & 0.89        \\ \hline
RNN   &   -          & 0.20   & 0.15   & 0.46        \\ \hline
mWDN   &   -         & 0.89   & -      & 0.87        \\ \hline
\end{tabular}
\label{tab:har}
\end{table}

\subsubsection{UCIHAR dataset}
We compared the effectiveness of our hypothesis against several well-known time series classification baselines with encoders based on LSTM, GRU, TCN, RNN, mWDN, and InceptionTime. %\footnote{We used the implementations from the python-based TSAI library~\cite{tsai}}. 
Table~\ref{tab:har} shows the consistent and higher performance of TS2Vec across different models. The degree of accuracy appeared to be mainly associated with the dataset in downstream classification rather than the choice of the dataset used for pre-training, e.g. using PAMAP2 data to pre-train the base model and classifying on UCIHAR achieved the same level of accuracy compared to using UCIHAR itself for pre-training. Even with a randomly initialised encoder model, the classification performance remains high for all three datasets. Among all the fully-supervised baselines, only InceptionTime was found to be effective; the others performed quite poorly or were so resource intensive to train (mWDN on PAMAP2). Overall, our proposed SSL-based pre-training has shown a positive impact on the downstream classifiers to learn new data. As it is shown, our proposed framework (TS2Vec with pretraining) consistently reaches comparably high scores across all datasets.
In the following, we explore two incremental setups on HAR datasets:

\begin{figure}
    \centering
    \includegraphics[width=0.95\linewidth]{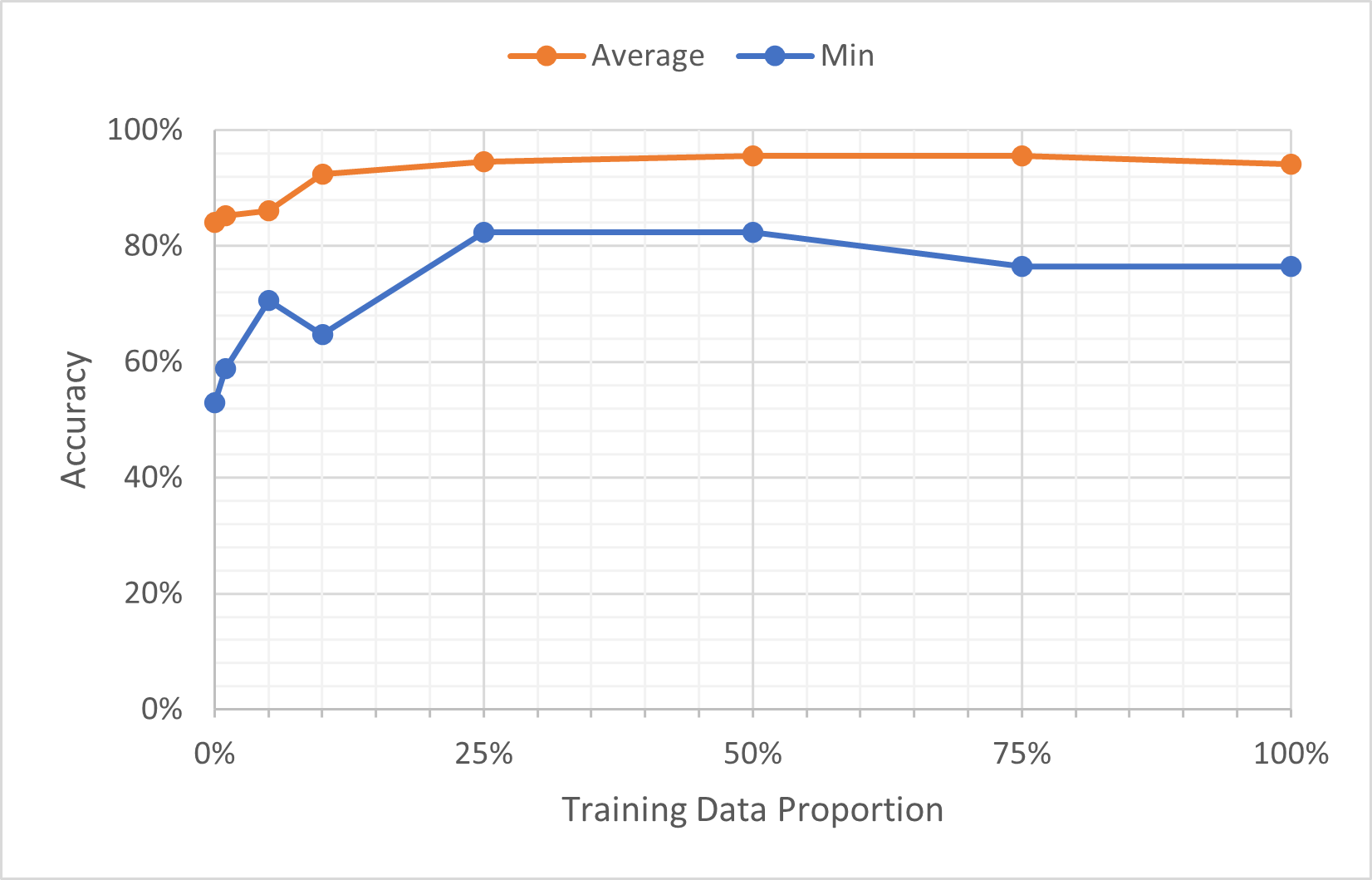}
    \caption{Instance-incremental setup for UCIHAR dataset. Performance has been improved over the first 25\% of training data but plateaued afterwards.}
    \label{fig:ucihar_ni}
\end{figure}

 \paragraph{Instance-incremental} As shown in Figure~\ref{fig:ucihar_ni}, increasing the proportion of data used in the pre-training phase improves the final accuracy. However, there is no noticeable improvement in accuracy achieved for the higher ratio of pre-training data (more than 25\% of the training data); this outcome also confirms the findings in~\cite{tstcc,gorade}.

\begin{figure}
    \centering
    \includegraphics[width=.95\linewidth]{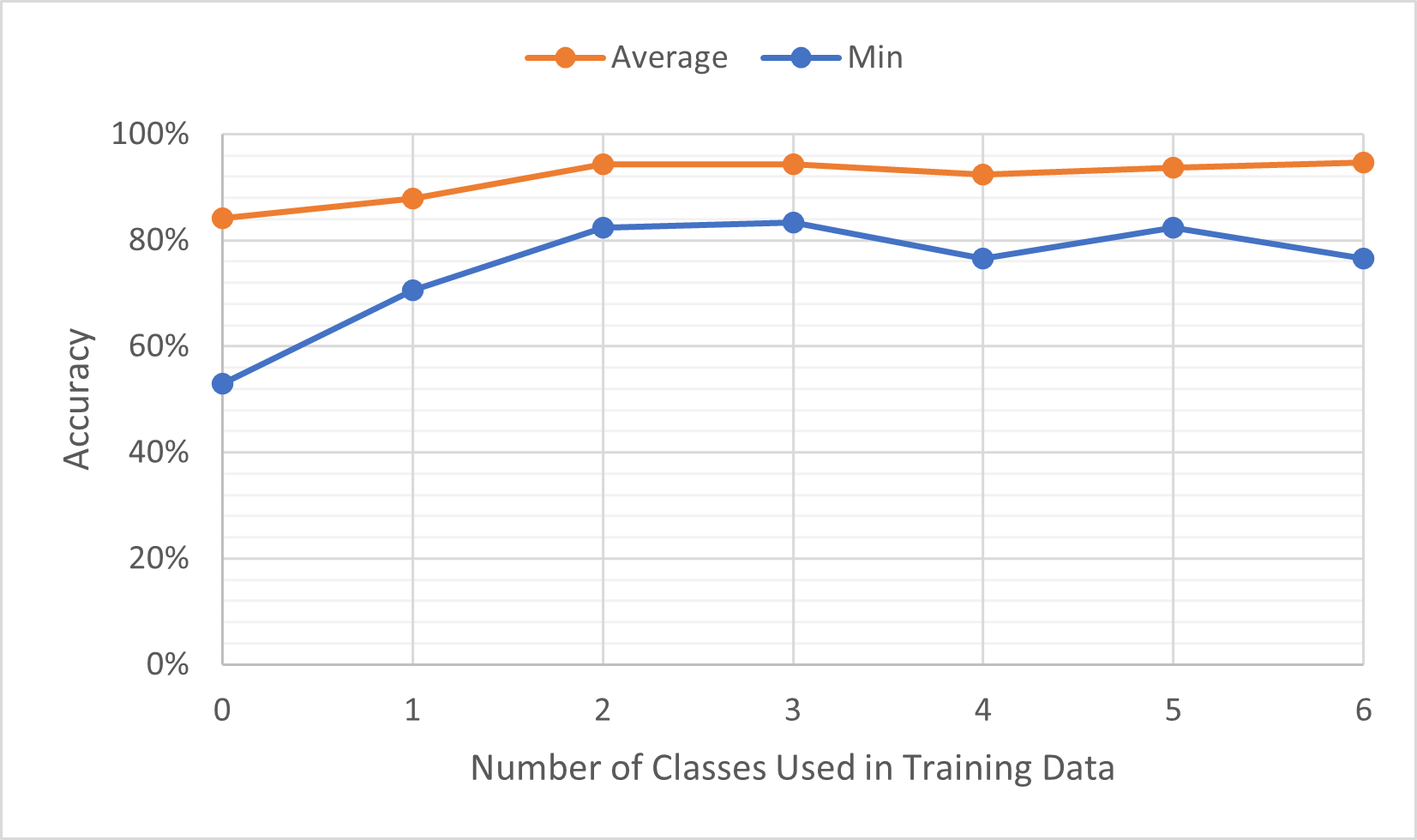}
    \caption{Class-incremental setup for UCIHAR dataset. Performance increased over the first two classes used for pre-training and plateaued afterwards.}
    \label{fig:ucihar_nc}
\end{figure}

\paragraph{Class-incremental} 
In this scenario, each class contains an approximately equal proportion of data. Increasing the number of classes utilized in the pre-training phase, starting from none to all 6 classes in the UCIHAR dataset, resulting in an improvement in downstream phase accuracy, peaking at two classes, as shown in Figure~\ref{fig:ucihar_nc}. There was a subtle difference in accuracy between an encoder model pre-trained with a subset of data containing instances from two of the classes compared to the full set with six classes. Since the classes were balanced, using 2 out of 6 classes equates to around 33\% of available training data, the accuracy achieved (94\%) was similar to the previous instance incremental scenario at 25\% proportion (95\%). That suggests that, for the UCIHAR dataset, the pre-training model has learned most of the features using only one-quarter of the data.

\begin{table*}
\caption{Accuracy on 30 UEA datasets in different scenarios.}
\resizebox{\textwidth}{!}{%
\begin{tabular}{|l|p{0.1\textwidth}|p{0.1\textwidth}|p{0.1\textwidth}|p{0.1\textwidth}|p{0.1\textwidth}|p{0.1\textwidth}|}
\hline
Dataset                   & Reported at \cite{ts2vec}     & Reproduced    & No data pre-training                  & Full Data pre-training                         & Random Subset pre-training               & Single Class Subset  pretrain      \\ \hline\hline
ArticularyWordRecognition & \textbf{0.99} & \textbf{0.99} & 0.98                        & \textbf{0.99}                        & 0.98                        & 0.95                        \\ \hline
AtrialFibrillation        & 0.20          & 0.20          & \textcolor{blue}{ 0.07} & 0.20                                 & 0.20                        & \textbf{0.22}               \\ \hline
BasicMotions              & \textbf{0.98} & \textbf{0.98} & 0.97                        & 0.97                                 & \textbf{0.98}               & 0.97                        \\ \hline
CharacterTrajectories     & \textbf{1.00} & 0.99          & 0.98                        & 0.99                                 & 0.99                        & 0.99                        \\ \hline
Cricket                   & 0.97          & \textbf{0.99} & \textbf{0.99}               & \textbf{0.99}                        & 0.97                        & 0.96                        \\ \hline
DuckDuckGeese             & \textbf{0.68} & 0.50          & 0.54                        & 0.46                                 & 0.48                        & 0.50                        \\ \hline
EigenWorms                & 0.85          & 0.83          & 0.82                        & 0.82                                 & \textbf{0.87}               & 0.85                        \\ \hline
Epilepsy                  & 0.96          & 0.96          & 0.95                        & 0.96                                 & \textbf{0.97}               & 0.96                        \\ \hline
ERing                     & 0.87          & 0.85          & \textbf{0.89}               & 0.85                                 & 0.85                        & 0.83                        \\ \hline
EthanolConcentration      & \textbf{0.31} & \textbf{0.31} & 0.27                        & 0.26                                 & 0.27                        & 0.28                        \\ \hline
FaceDetection             & 0.50          & \textbf{0.51} & \textbf{0.51}               & \textbf{0.51}                        & \textbf{0.51}               & \textbf{0.51}               \\ \hline
FingerMovements           & 0.48          & 0.50          & \textcolor{blue}{ 0.45} & \textcolor{red}{ \textbf{0.56}} & 0.52                        & \textcolor{red}{ 0.55} \\ \hline
HandMovementDirection     & 0.34          & 0.31          & 0.27                        &  \textcolor{red}{\textbf{0.36}} & 0.31                        & 0.29                        \\ \hline
Handwriting               & 0.52          & \textbf{0.55} & \textcolor{blue}{ 0.25} & \textbf{0.55}                        & \textcolor{blue}{ 0.46} & \textcolor{blue}{ 0.43} \\ \hline
Heartbeat                 & 0.68          & 0.69          & \textbf{0.72}               & 0.69                                 & 0.71                        & 0.70                        \\ \hline
InsectWingbeat            & \textbf{0.47} & 0.46          & \textbf{0.47}               & \textbf{0.47}                        & 0.46                        & \textbf{0.47}               \\ \hline
JapaneseVowels            & \textbf{0.98} & \textbf{0.98} & 0.97                        & \textbf{0.98}                        & \textbf{0.98}               & \textbf{0.98}               \\ \hline
Libras                    & \textbf{0.87} & 0.84          & 0.83                        & 0.84                                 & 0.86                        & 0.81                        \\ \hline
LSST                      & 0.54          & 0.55          & \textbf{0.59}               & 0.57                                 & 0.56                        & 0.57                        \\ \hline
MotorImagery              & \textbf{0.51} & 0.50          & 0.50                        & 0.50                                 & 0.50                        & 0.46                        \\ \hline
NATOPS                    & 0.93          & 0.91          & 0.93                        & 0.92                                 & \textbf{0.94}               & 0.91                        \\ \hline
PEMS-SF                   & \textbf{0.68} & 0.65          & 0.65                        & 0.66                                 & 0.66                        & \textbf{0.68}               \\ \hline
PenDigits                 & \textbf{0.99} & \textbf{0.99} & 0.98                        & \textbf{0.99}                        & \textbf{0.99}               & \textbf{0.99}               \\ \hline
PhonemeSpectra            & 0.23          & 0.23          & 0.21                        & 0.23                                 & \textbf{0.24}               & \textbf{0.24}               \\ \hline
RacketSports              & \textbf{0.86} & \textbf{0.86} & 0.77                        & \textbf{0.86}                        & \textbf{0.86}               & 0.83                        \\ \hline
SelfRegulationSCP1        & \textbf{0.81} & 0.77          & 0.78                        & 0.79                                 & 0.77                        & 0.80                        \\ \hline
SelfRegulationSCP2        & \textbf{0.58} & 0.55          & 0.57                        & \textbf{0.58}                        & \textbf{0.58}               & 0.56                        \\ \hline
SpokenArabicDigits        & \textbf{0.99} & \textbf{0.99} & \textcolor{blue}{ 0.92} & 0.97                                 & 0.97                        & 0.97                        \\ \hline
StandWalkJump             & 0.47          & 0.47          & \textcolor{blue}{ 0.27} & \textcolor{red}{ \textbf{0.53}} & \textcolor{blue}{ 0.42} & \textcolor{blue}{ 0.33} \\ \hline
UWaveGestureLibrary       & \textbf{0.91} & 0.90          & \textcolor{blue}{ 0.69} & 0.90                                 & 0.90                        & \textcolor{blue}{ 0.85} \\ \hline\hline
Average                   & \textbf{0.70} & 0.69          & 0.66                        & \textbf{0.70}                        & 0.69                        & 0.68                        \\ \hline
\end{tabular}%
}
    \label{tab:uea}
\end{table*}

\subsubsection{UEA dataset}
We were able to verify the effectiveness of TS2Vec on UEA datasets by comparing the reported scores in column (2) and reproduced scores in column (3) of Table~\ref{tab:uea}. This table also showed that in the process of adapting the TS2Vec model for experimentation in our workflow, we did not significantly modify the behaviour or performance of the original, as seen from the scores in column (5) compared to columns (2) and (3).
Using paired t-test between columns for these 30 datasets, we found a significant difference between pre-training without data and with full training data (p-value 0.02), but no statistical difference between pre-training with full data and a subset of randomly sampled data or data from a single class (p-value greater than 0.05). This suggests that while using a representation learning model and pre-training does increase classification accuracy, applying pre-training on even a small subset of data can achieve performance like that of the whole dataset. It is also interesting that pre-training on a subset of data that contains only instances from a single class does not seem to deteriorate performance significantly.

There were cases where the random subset pre-training or the single class pre-training produced noticeably lower scores (more than 5\% decrease) than the reproduced score, but most of these had low (below 60\%) scores even with full data used for pre-training. Hence, for many of these datasets, the choice of data for pre-training does not make much difference.

\section{Discussions and Limitations}
The key advantage of self-supervised learning is about the capability of learning useful representation from unannotated data. With easily obtainable unlabeled data, it becomes natural to incrementally incorporate new information as soon as it becomes available for the self-supervised learning models.
The study of self-supervised learning with incremental data has the potential to provide advanced capabilities for real-world applications in which data is constantly changing and new information becomes available.
Hence, this paper presents an empirical study on the self-supervised learning paradigm using incremental time series data.

Given that this work is the first empirical study of continuous learning using self-supervised learning in time-series and wearable data where data is presented incrementally, there are several limitations that we aim to address in future work. First, as explained in the earlier section, we considered the incremental presence of data in the pretraining phase. Then we transferred the knowledge to train the downstream task (i.e., classifier) based on a fully available dataset. However, in real-world situations, the data is continuously generated, and this should be considered in both upstream pretraining and downstream tasks.
Along with the main motivation of the self-supervised learning paradigm, we will investigate the impact of incremental SSL techniques in low-labelled data regimes.

\section{Conclusion}
In this study, we were interested in examining whether a time series classification problem can leverage the vast amount of unlabelled data that becomes incrementally available during the pre-training phase under the self-supervised learning paradigm.
To this end, we evaluated the impact of self-supervised pre-training on the time series classification task using data from various sources, including different distributions and domains.
We also examined how the performance of the classification task changes as more data is made available during pre-training, simulating scenarios where all of the data is not immediately accessible. Additionally, we explored the impact of training the encoder and classifier based on data from different domains and distributions.
Based on our experimental results on multiple datasets, we observed that the pre-training stage did not have a noticeable benefit from more pre-training data and the classification performance was not improved. Conversely, when the pre-training was performed only on a subset of classes or when data was partitioned to form a different distribution, the final performance did not deteriorate. Even when pre-training and classification were done on datasets from different domains, the performance was comparable to other time series classification models.

\section*{Acknowledgment} 
The authors would like to thank Dr Dong Gong for their valuable contributions to this work. Also, they acknowledge the support of the Artificial Intelligence for Decision Making (AI4DM) research initiative grant from the Defence Science \& Technology Group for the project titled ``Online learning-based forecasting with irregular time-series data''.

% the end of the paper, preceded by an unnumbered run-in heading (i.e.
% 3rd-level heading).

\bibliographystyle{splncs04}
\bibliography{main}

\end{document}